\documentclass[dvipsnames]{article}
\usepackage{ijcai17}

\usepackage{times}

\usepackage{amsmath}
\usepackage{amssymb}
\usepackage[plain,Figure]{algorithm}
\usepackage[noend]{algpseudocode}
\usepackage{graphicx}
\usepackage{stmaryrd}
\usepackage{mathtools}
\usepackage{xspace}
\usepackage{color}
\usepackage{comment}
\usepackage{multirow}
\usepackage{amsthm}
\usepackage{array}
\usepackage{booktabs}
\usepackage{pifont}
\usepackage{wrapfig}
\usepackage{amsthm}
\usepackage{pgfplotstable}
\usepackage{caption}
\usepackage{program}
\usepackage{lscape}

\pgfplotstableset{
    every head row/.style={
    before row=\toprule,after row=\midrule},
    every last row/.style={
    after row=\bottomrule},
    col sep = &,
    row sep=\\,
    string type,
}

\begin{document}

\setlength{\pdfpageheight}{\paperheight}
\setlength{\pdfpagewidth}{\paperwidth}

\renewcommand{\algorithmiccomment}[1]{ {\itshape \texttt{//} #1}}
\newcommand{\spec}{\phi}

\algnewcommand\algorithmicforeach{\textbf{foreach}}
\algdef{S}[FOR]{ForEach}[1]{\algorithmicforeach\ #1\ \algorithmicdo}
\renewcommand{\t}[1]{\texttt{#1}}
\newcommand{\sem}[1]{\llbracket #1 \rrbracket_{v}}
\newcommand{\sems}[1]{\llbracket #1 \rrbracket_{v}}

\newcommand{\eg}{\textit{e.g.}}
\newcommand{\ie}{\textit{i.e.}}
\newcommand{\etc}{\textit{etc.}}
\newcommand{\figref}[1]{Figure~\ref{#1}}
\newcommand{\namecite}[1]{\citeauthor{#1}~(\citeyear{#1})}

\theoremstyle{definition}
\newtheorem{example}{Example}
\newcommand{\quotes}[1]{``#1''}

\newcommand{\lookupapi}{\mathcal{L}}
\newcommand{\regexapi}{\mathcal{R}}
\newcommand{\transformapi}{\mathcal{T}}
\newcommand{\exref}[1]{Example~\ref{#1}}
\newcommand{\tool}{\textsc{Dapip}}

\renewcommand{\ttdefault}{cmss}


\title{Deep API Programmer: Learning to Program with APIs}

\author{Surya Bhupatiraju \\
MIT \\
surya@mit.edu \And
Rishabh Singh \\
Microsoft Research \\
risin@microsoft.com \And
Abdel-rahman Mohamed \\
Microsoft Research \\
asamir@microsoft.com \And
Pushmeet Kohli \\
Microsoft Research \\
pkohli@microsoft.com}

\maketitle

\begin{abstract}
We present $\tool$, a Programming-By-Example system that learns to program with APIs to perform data transformation tasks. We design a domain-specific language (DSL) that allows for arbitrary concatenations of API outputs and constant strings. The DSL consists of three family of APIs: regular expression-based APIs, lookup APIs, and transformation APIs. We then present a novel neural synthesis algorithm to search for programs in the DSL that are consistent with a given set of examples. The search algorithm uses recently introduced neural architectures to encode input-output examples and to model the program search in the DSL. We show that synthesis algorithm outperforms baseline methods for synthesizing programs on both synthetic and  real-world benchmarks.
\end{abstract}


\section{Introduction}

The ability to discover a program consistent with a given user intent (specification) is considered as one of the central problems in artificial intelligence~\cite{greensynthesis}. While significant progress has been made in synthesizing programs in different domains~\cite{sygus}, current synthesis techniques do not scale to larger and more complex programs. Moreover, the state-of-the-art synthesis techniques~\cite{cacm12} require a great deal of domain expertise with manually designed heuristics and rules to develop an efficient search procedure. In this paper, we present $\tool$, or Deep API Programmer, a system that aims to overcome some of these shortcomings by automatically learning a synthesis algorithm in the domain of data transformation tasks.

The process of transforming data from raw data into usable formats (also known as \emph{data wrangling}) is a key problem faced by data scientists for any data analysis task. Some studies have reported that this data wrangling process can sometimes take up to 80\% of the total data analysis time~\cite{dasu2003exploratory,wrangler}. Recently, Programming-By-Example (PBE) techniques such as FlashFill~\cite{flashfill,cacm12} and BlinkFill~\cite{blinkfill} were developed to help users perform data transformation tasks using examples instead of having to write complex programs. These techniques encode the space of programs using a domain-specific language (DSL), and then develop algorithms based on version-space algebra (VSA)~\cite{flashmeta,lauvsa} to efficiently search the space of programs. There are two key shortcomings of these approaches. First, the DSL is limited to only certain low-level \emph{syntactic} regular expression-based operators that allow for an efficient structuring of search space. This limits the expressiveness of the PBE systems; for example, they do not allow \emph{semantic} data transformations using arbitrary transformation functions such as obtaining month names from a date or abbreviating the state name in an input address. Second, building an efficient synthesizer using VSA requires a large engineering effort with manually designed heuristic rules.

We tackle the first shortcoming by designing $\tool$'s DSL to have function APIs as the core element, which allows for composition of APIs with constant strings. The DSL consists of three kinds of APIs: regular expression-based APIs, lookup APIs, and transformation APIs. The regular expression-based APIs perform a regular expression-based transformation on the input strings, which are needed for syntactic data transformations. The lookup APIs search for a particular string in the input data based on a dictionary of strings, and the transformation APIs perform some transformation on top of a lookup operation based on a predefined mapping between two sets of strings. The lookup and transformation APIs allow for semantic data transformations.

The second shortcoming is handled by learning the synthesis algorithm in $\tool$ automatically from data using two recently introduced neural modules~\cite{neuralflashfill}. The first module called the \emph{cross-correlational encoder} computes a fixed-dimension vector representation of the input-output examples by using tensor representations obtained by running two bi-directional LSTMs~\cite{hochreiter1997long,bilstm} on the input and output strings and computing their cross correlation. The second module, the recursive-reverse-recursive neural network, or {\it R3NN}, encodes a partial derivation in the DSL and given the example encoding vector, returns a distribution over the space of possible expansions to the partial derivation. The R3NN incrementally builds a program in the DSL that is consistent with the input-output examples. The input-output encoder and the R3NN modules are trained end-to-end using thousands of programs and corresponding input-output examples, which are automatically sampled from the DSL.

We evaluate $\tool$ on a set of synthetic and 238 real-world FlashFill benchmarks. Our experiments indicate that our deep learning based approach is able to effectively model and predict the presence of different types of APIs. It is able to solve 45\% of the FlashFill benchmarks and significantly outperforms the enumerative search based baseline.

To summarize, the key contributions of this paper are:
\begin{itemize}
\item We design an expressive DSL with APIs that can encode both syntactic and semantic data transformation tasks.
\item We automatically learn a synthesis algorithm for synthesizing programs in the DSL using neural architectures.
\item We evaluate our system $\tool$ on 238 real-world FlashFill benchmarks and thousands of synthetic benchmarks.
\end{itemize}

\section{Motivating Examples}
We present a few real-world examples to motivate the DSL.
\begin{example}
An Excel user wanted to transform names to first initial followed by last name as shown in \figref{flashfilltask}. Since some input examples had optional middle names, the user was struggling to find a macro to perform the desired task.
\end{example}

\begin{figure}[t]
\begin{center}
\begin{tabular}{|c|c|c|}
\hline
& \multicolumn{1}{|c|}{\bf Input $v$} & \multicolumn{1}{|c|}{\bf Output} \\\hline\hline
1 & John S. Henry & {J. Henry} \\ \hline
2 & Mike Stanley & {M. Stanley} \\ \hline
3 & Bernie John Smith & {\bf B. Smith} \\ \hline
4 & Martha S Johnson & {\bf M. Johnson} \\ \hline
\end{tabular}
\end{center}
\caption{An example FlashFill task of abbreviating names. The user provides the first two outputs, and the bold entries are then automatically generated by the learned program.}
\label{flashfilltask}
\end{figure}


$\tool$ learns the following program for this task: $\t{Concat}(\t{GetFirstChar}(v),\t{ConstStr}(\quotes{.}),\t{GetLastWord}(v))$. The learned program uses the \t{GetFirstChar} and \t{GetLastWord} APIs that belong to the class of regex APIs, which extract substrings from the input string based on regular expressions.


\begin{figure}[t]
\begin{center}
\begin{tabular}{|c|c|c|}
\hline
& \multicolumn{1}{|c|}{\bf Input $v$} & \multicolumn{1}{|c|}{\bf Output} \\\hline\hline
1 & 500 Mem Dr., Cambridge, 02139 & {Cambridge, MA} \\ \hline
2 & 22 NE Street, Redmond, USA & {Redmond, WA} \\ \hline
3 & Seattle, 98002 & {\bf Seattle, WA} \\ \hline
4 & 21 Peace Ave., Kirkland, 98034 & {\bf Kirkland, WA} \\ \hline
\end{tabular}
\end{center}
\caption{Transforming addresses to City and State.}
\label{semanticapi}
\end{figure}

\begin{example}
An Excel user had a list of addresses and wanted to extract the city and state values as shown in \figref{semanticapi}.
\end{example}

This is an example of a very common task that can not be performed by systems such as FlashFill. Since the data is in many different formats, there is no consistent regular expression that can be used to extract the city names. Moreover, to obtain the state name, the system needs to use a transform API \t{GetStateFromCity}. $\tool$ learns the following program:$\t{Concat}(\t{GetCity}(v),\t{ConstStr}(\quotes{,}),$\\$\t{GetStateFromCity(GetCity}(v)))$.

More examples of real-world FlashFill tasks can be found in Appendices~\ref{ff_samples} and~\ref{ff_unsolved}.

\begin{figure}[t]
\includegraphics[scale=0.4]{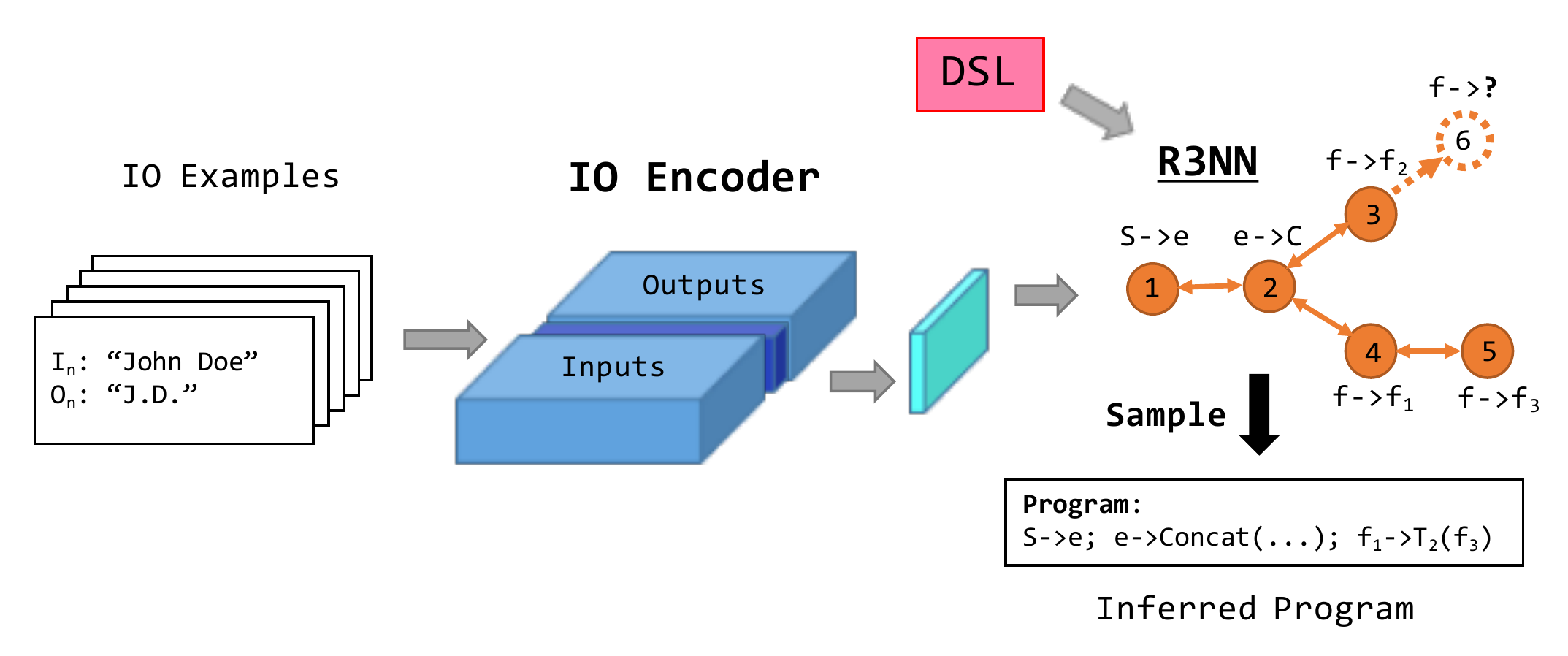}
\caption{Using the trained R3NN model to sample programs from the DSL given a set of input-output examples; even in the inference use case, nodes are expanded in a particular, discrete order.}
\label{testoverview}
\end{figure}

\section{Overview of Approach}

\begin{figure*}[h]
\centering
\includegraphics[scale=0.5]{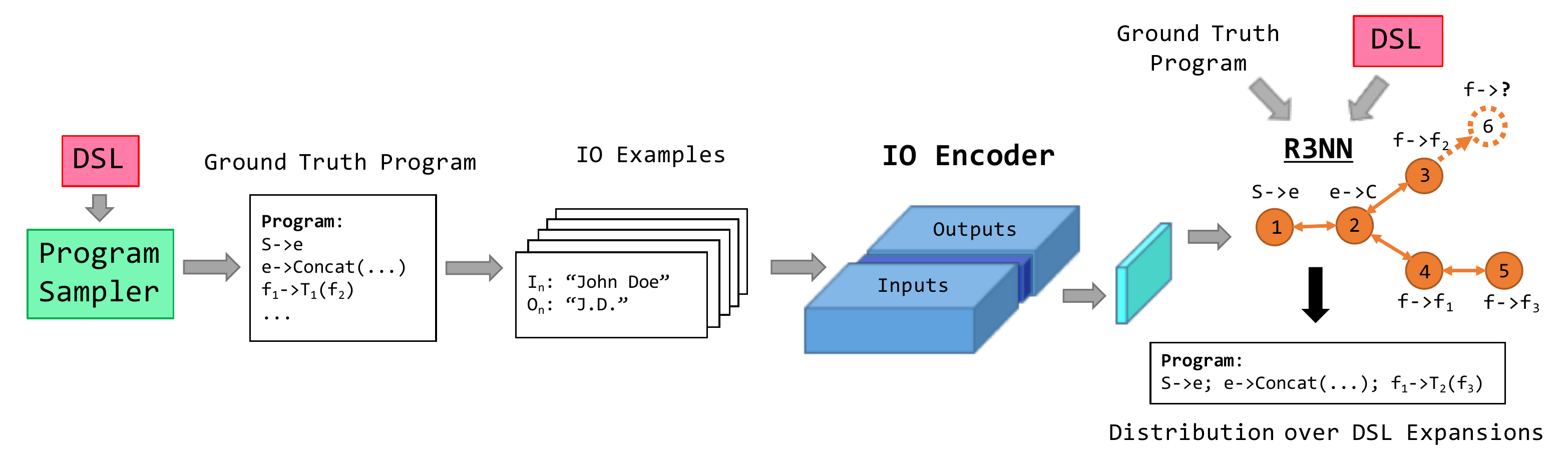}
\caption{Training the R3NN network to learn distributions over DSL expansions conditioned on the input-output examples; expansion are performed in a particular order as dictated by the conditional distribution.}
\label{trainingoverview}
\end{figure*}

We now present an overview of our end-to-end system that learns to synthesize programs in a DSL that are consistent with a set of examples. The training phase of our system is shown in \figref{trainingoverview} and the test phase is shown in \figref{testoverview}. We first design a DSL that allows for composition of nested API calls with constant strings. We designed this DSL after studying a large family of real-world string transformation tasks so that it is expressive enough to encode these tasks. During the training phase, we use a program sampler to uniformly sample a large number of programs from this DSL. For each program, we use a rule-based approach to construct 5 input strings for the program such that the prerequisites of the program are met. We obtain the output strings by executing the program on the input strings.

During training, each sampled program together with the corresponding input-output examples is used to train the \emph{R3NN} model, a neural architecture that learns distributions over the expansions in the DSL conditioned on the examples. The examples are encoded using a second neural architecture called the \emph{cross-correlational encoder}, which produces a fixed-dimensional vector. The R3NN system takes as input the input-output conditioning vector, the DSL, and the training program, and is trained to predict a conditional distribution over the set of DSL expansions. The next expansion is sampled from this conditional distribution, leading to the partial tree, and the procedure repeats; one can observe a potential order of the nodes growing in the respective figures.

The trained R3NN model can then be used to synthesize programs in the DSL given a set of examples. The trained model takes the input-output conditioning vector as input, and generates a distribution over the set of DSL expansions that are likely to be the expansions required to construct the desired program. The distribution is then sampled to derive programs in the DSL, where the order of expansions is specified by the distribution, as shown in the respective figure, and the system returns the first program that is consistent with the input-output examples.

\section{Domain-Specific Language}

%
%

The syntax of the domain-specific language for API-based string transformations is shown in \figref{dsl}. The top-level construct of the language is the \t{Concat} function that returns the concatenation of its argument substrings $f_i$. A substring expression $f$ can either be a constant string $s$, the input string $v$, or the result of an API function with $f$ as its argument. The \t{Concat} operator allows for composition of API calls with constant strings. The DSL consists of 3 types of APIs: regex APIs $\regexapi$, lookup APIs $\lookupapi$, and transformation APIs $\transformapi$.

\textbf{Regex API $\regexapi$: }
The regex APIs search for certain regular expression-based patterns in the input string and return the matched string. Some examples of regex APIs are \t{GetFirstNum}, \t{GetBetFirstAndSecondCommas}, etc. Our DSL consists of 104 such regex APIs.

\textbf{Lookup API $\lookupapi$: }
The lookup APIs look for presence of certain strings in the input string and return the lookup string. Each lookup API consists of a dictionary of a finite collection of strings, which are used for searching input substrings. Some examples of lookup APIs are \t{GetCity},\t{GetState}, \t{GetStockSymbol} etc. For example, the \t{GetState} API contains a dictionary of $50$ US state names, whereas the \t{GetCity} API contains a dictionary of $18,200$ US cities. Our DSL consists of $18$ such lookup APIs.

\textbf{Transformation API $\transformapi$: }
The transformation APIs consists of a dictionary $D:S_1 \rightarrow S_2$, which maps a finite collection of strings $S_1$ to another finite collection of strings $S_2$. These APIs search for a string $s \in S_1$ in the input string and return the corresponding output string $D[s] \in S_2$. Some examples of such APIs include \t{GetStateFromCity}, \t{GetFirstNameInitial}, etc. For example, the transformation API \t{GetStateFromCity} consists of a dictionary mapping a collection of $18,200$ US cities to the corresponding US states. Our DSL consists of $13$ such functions.

The full list of all functions is provided in Appendix~\ref{apis}.
%

\begin{figure}
\begin{eqnarray*}
\mbox{String } e & := & \t{Concat}(f_1, \ldots, f_n) \nonumber \\
\mbox{Substring } f & := & \regexapi(f) \; \; | \; \; \transformapi(f) \; \; | \; \; \lookupapi(v)\\
& | & \t{ConstStr}(s) \; \; | \; \; v \\
\mbox{Lookup API } \lookupapi & := & \lookupapi_1 \; \; | \; \; \cdots \; \; | \; \; \lookupapi_m \\
\mbox{Regex API } \regexapi & := & \regexapi_1 \; \; | \; \; \cdots \; \; | \; \; \regexapi_l \\
\mbox{Transform API } \transformapi & := & \transformapi_1 \; \; | \; \; \cdots \; \; | \; \; \transformapi_k
\end{eqnarray*}
\caption{The syntax of the DSL for API compositions.}
\label{dsl}
\end{figure}

\section{Neural Architecture for Search}
The neural search over the programs in the DSL conditioned on the input-output examples is performed using the model outlined in~\cite{neuralflashfill}. First, the input-output examples are encoded into a fixed length feature vector that aims to capturing shared patterns between the input and output strings. This example representation is then passed to a neural tree-based generative model over program trees, called R3NN, to generate the desired hidden program. We provide a high level overview of the both the architectures.

\subsection{Neural Input-Output Encoder}

The cross-correlational encoder generates a fixed-dimensional vector representation of a set of input-output (I/O) examples. Intuitively, the encoder needs to capture three key information: parts of the output strings that are likely to be constant strings, parts of the output strings that can be computed from input strings, and some characteristics of the example strings that will help the program generator module identify the set of useful APIs for the given task. To simplify the DSL, we assume a fixed universe of possible constant strings so that we can focus on training the encoder to produce the likely set of APIs.

The I/O encoder first runs two bidirectional LSTM networks separately on the input and output strings in each example pair, which produces two matrices of size $2\times H\times T$, where $H$ is the LSTM hidden dimension and $T$ is the maximum length of the I/O string. The encoder then slides the output matrix over the input matrix for each time step and computes the outer product between respective matrix columns. There are in total $2\cdot(T-1)$ alignments as we slide the matrices and we obtain $2\cdot(T-1)$ vectors in total after the dot product. Finally, the encoder concatenates the values for overlapping time steps to obtain a $2\times T\times (T-1)$-dimensional vector encoding for each example pair.

%
%

\subsection{Tree-Structured Generation Model}

The tree generation model incrementally constructs a program tree starting from the start symbol of the DSL grammar and expanding the tree with one derivation at a time until obtaining a tree with that consists only non-terminal nodes. The R3NN network assigns posterior probabilities to every valid expansion of a partial tree to guide the search algorithm. In other words, given a partial program tree, the R3NN network decides which non-terminal node to expand in the tree and with which expansion rule in the grammar.

The R3NN is defined by the following parameters: i) an $M$-dimensional representation $\phi(s) \in \mathbb{R}^M$ for every symbol $s$ in the grammar, ii) an $M$-dimensional representation $\omega(r) \in \mathbb{R}^M$ for each grammar rule $r$, iii) a deep neural network $f_r$ for each grammar rule $r$ that takes as input a vector $x \in \mathbb{R}^{Q \cdot M}$ (where Q is the number of RHS symbols of $r$) and outputs a vector $y \in \mathbb{R}^M$, and iv) a deep neural network $g_r$ (reverse of $f_r$) which takes as input a vector $x' \in \mathbb{R}^{Q \cdot M}$ and outputs a vector $y' \in \mathbb{R}^{Q \cdot M}$.

Given a partial program tree, R3NN first assigns the representation $\phi(S(l))$ to each leaf node $l$, where $S(l)$ denotes the grammar symbol of node $l$. It then performs a standard recursive pass over the tree from bottom-to-top, by recursively applying $f_{R(n)}$ for every non-leaf node $n$ on its RHS node representations to compute the representation of $n$, where $R(n)$ denotes the rule associated with node $n$. This pass continues until we reach the root node. The $\phi(root)$ represents information about all tree nodes, but does not encode any notion of the node positions in the tree. To solve this issue, R3NN performs a reverse-recursive pass starting from the root node to compute updated representations of all child nodes using the reverse deep network $g_{R(n)}$. After performing the reverse-recursive pass, each leaf node $l$ is assigned a new distributed representation $\phi'(l)$, which intuitively captures the global information about every other node in the tree.

The scores for each expansion $e \in E$ can now be obtained from the global leaf representations $\phi'(l)$. Let $e_r$ be the expansion type (production rule $r \in R$ that $e$ applies) and let $e_l$ be the leaf node $l$ that $e_r$ is applied to for an expansion $e$. The score of an expansion is calculated using $z_e = \phi'(e_l) \cdot \omega(e_r)$ and the probability of the expansion is obtained by exponentiated normalized sum over the scores: $\pi(e) = \frac{e^{z_e}}{\sum_{e' \in E} e^{z_{e'}}}$.

\section{Evaluation}

We now present results from two major sets of experiments and analyze the model in more detail with the goal of assessing its expressiveness. We demonstrate that our model is capable of learning to synthesize simple programs when provided with a library of over 100 API functions. We also show that the model is capable of \textit{strong generalization}, where it can not only generalize across different I/O examples for a given program, but also across new, unseen programs.

\subsection{Experimental Setup and Training Details}
\label{expsetup}
%

We use both synthetic benchmarks and real-world FlashFill benchmarks for evaluation. The synthetic benchmarks are obtained by sampling the programs in the DSL uniformly, and then using a rule-based approach to generate corresponding input-output examples. For example, if we sample a program consisting of \t{GetThirdNum} and \t{GetState} APIs, the rule-based approach would ensure that the input strings in the example consist of at least three numbers and one state strings. For each benchmark, we sample five input strings and the corresponding output strings are obtained by executing the sampled program on the input strings. Several examples of training data are shown in Appendix~\ref{ffp_samples}.

We first train the R3NN on a DSL consisting of only one family of APIs to evaluate its effectiveness on learning individual API family. We call the models trained on only the regex APIs (and constant strings) as the \t{FF} models and call the corresponding DSL as the {\it regex-only DSL}. We then train the R3NN with all APIs to evaluate the effectiveness of learning programs in the DSL consisting of different APIs and their composition with constant strings; we call this DSL the {\it full DSL}. The models trained on the full DSL are called the \t{FF++} models. Since the FlashFill benchmarks can be solved using only the regex APIs and the set of constant strings, we also evaluate the \t{FF} model on the FlashFill benchmarks.

We train the cross-correlation encoder and R3NN jointly with the principle of maximum likelihood; the model produces posterior probabilities over possible expansions and we backpropagate an error signal based on the ground truth programs. We use the Adam optimizer~\cite{kingma2014adam}, with an initial learning rate of 0.001 and clipping gradients at 10 for both modules. We found that small learning rates are crucial for R3NN to prevent unstable learning. Every epoch consists of 1000 training batches of 10 instances, where each instance contains a ground truth program and 5 input-output pairs. The evaluation on synthetic data is performed on programs that are not seen during training. We report results when evaluating with both 1-best inference and with stochastic search (10, 50, or 100 samples), where we resample a program conditioned on the same input-output examples multiple times. This way, we allow the model to have small errors in its final posterior probabilities for selecting an expansion.

\subsection{Learning API Types}
Each of the three classes of API functions, while much more interpretable, still pose nontrivial challenges for the model to learn to compose. The lookup API functions contain large dictionaries and the model must learn when to call such APIs given the input-output examples. For example, while the difference between names and cities may seem trivial to human practitioners, the model must learn to disambiguate each of these entities. The transformation API functions pose an additional challenge; with programs that require these types of API calls, not only does the model need to learn some encoding of the hidden dictionary, but the output string may not contain any obvious matching substring in the input string because of the nature of the API function. As a result, a simple string matching algorithm between the inputs and outputs will not work to solve this problem, and the input-output encoder must learn useful representations of pairs of them, and be expressive enough to capture the implicit string transformation. Lastly, the regex API functions do not encode dictionaries but represent syntactic substring operations, and the model must learn to recognize which API functions to call based on which parts of the output are present in the input.

We first present an ablative study of what class of APIs are the easiest to learn in isolation, and which one is the most challenging in the full DSL.

\paragraph{Regex APIs}

In Table \ref{tab:ff1}, we report the training and validation set accuracies of different models trained on the regex-only DSL (\t{FF} model). The length column denotes the maximum length of programs that each model was trained on. The length 7 model was trained with 9000 programs, length 8 with 16000, length 9 with 616510, and length 10 with 1263000 programs. For validation, we select 1000 randomly chosen held-out programs from this set and generate new I/O examples to test the generalization power of the trained model.

\begin{table}[h]
\centering
\pgfplotstabletypeset{
Length & Training & Validation \\
7 & 80\% & 67\% \\
8 & 91\% & 85\% \\
9 & 22\% & 20\% \\
10 & 64\% & 63\% \\
}
\captionof{table}{Best \t{FF} model performance by max program length.}\label{tab:ff1}
\vspace{-0.25mm}
\end{table}

Of particular note is the performance on programs of length 10. At this length, the DSL can generate programs with API nesting, API composition, and concatenation with a constant string; this represents all possible constructs in our DSL.

\paragraph{Lookup and Transform APIs}

In this experiment, we fix the maximum size of the programs in the training and validation set to size 10 and only include the lookup and transform APIs in the DSL. The results are shown in Table \ref{tab:abl2}. We find that when the DSL is restricted to these APIs, the trained models achieve a very high accuracy and are able to identify composition of APIs with very high precision.

\begin{table}[h]
\centering
\pgfplotstabletypeset{
APIs & Training & Validation \\
Lookup & 96\% & 96\% \\
Lookup+Transform & 98\% & 98\%  \\
}
\captionof{table}{Learnability of other APIs.}\label{tab:abl2}
\end{table}

\paragraph{All APIs: Regex + Transform + Lookup}

We now present the model evaluation that was trained on the full DSL. Recall that because we've trained on the full DSL, these models are referred to as the \t{FF++} models.

\begin{table}[h]
\centering
\pgfplotstabletypeset{
Length & Training & Validation \\
7 & 54\% & 46\%  \\
8 & 75\% & 64\%   \\
9 & 46\% & 37\%  \\
10 & 50\% & 44\%  \\
}
\captionof{table}{The performance of best \t{FF++} model on synthetic dataset by max length of programs.}\label{tab:ffp1}
\end{table}

The performance of the \t{FF++} models is shown in Table \ref{tab:ffp1}. We observe that both training and validation accuracies decreased as compared to the \t{FF} models, which is expected since we now have an increased set of APIs that also include more complex APIs encoding large dictionaries. However, the length 10 model is still able to get 44\% accuracy.

We analyze these results further to understand the learnability of different APIs when trained together as shown in Table \ref{tab:abl1}. The regex APIs seem to be the easiest to learn for the network, which may be accredited to the specific nature of the IO encoder, as it was designed to detect patterns in substrings between the input and output examples. Interestingly, the lookup APIs are harder to learn than the transformation APIs, which can be attributed to the fact that they encode larger dictionaries as compared to the dictionaries of transform APIs.

\begin{table}[h]
\centering
\pgfplotstabletypeset{
Samples & LookupAPIs & TransformAPIs & RegexAPIs \\
10 & 32\% & 50\% & 72\% \\
50 & 37\% & 52\% & 89\% \\
100 & \textbf{38\%} & \textbf{57\%} & \textbf{92\%} \\
}
\captionof{table}{Ablative analysis of \t{FF++} model performance.}\label{tab:abl1}
\end{table}

\subsection{FlashFill using API Compositions}\label{results}
We now present the results of the best \t{FF} and \t{FF++} models on the FlashFill benchmarks obtained from the authors of FlashFill~\cite{cacm12}. These benchmarks correspond to real-world string transformation tasks in Excel, where each benchmark comprises of 5 input-output string examples.

\subsubsection{\t{FF} models}

\paragraph{Baseline performance with uniform search}
We first present the results we obtain with a baseline uniform search model on the FlashFill benchmarks in Table~\ref{tab:ffu}. The baseline model performs a uniform search over the DSL expansions and is biased towards small programs. We also present stochastic sampling results for a fair comparison with the performance of the \t{FF} models.

\begin{table}[h]
\centering
\pgfplotstabletypeset{
Samples & 10 & 50 & 100 \\
Performance & 2\% & 10\% & 17\% \\
}
\captionof{table}{Uniform search on FF benchmarks}\label{tab:ffu}
\vspace{-0.2cm}
\end{table}

The uniform search does surprisingly well considering the large space of all possible programs because the DSL we designed with APIs allows many of the benchmarks to be solved with a single call, e.g. \texttt{GetFirstWord}, and the uniform search sampler is biased towards shorter programs.

\paragraph{\t{FF} Model performance on FlashFill Benchmarks} We now evaluate the trained models whose accuracies on synthetic data are reported in Table \ref{tab:ff1}. Note that unlike in Table \ref{tab:ff1}, each of the model is evaluated on the same dataset and so the results are comparable across rows. In this case, we not only report the results with stochastic sampling, but also report the 1-best programs under the 1 column in Table \ref{tab:ff2}.

\begin{table}[h]
\centering
\pgfplotstabletypeset{
Length & 1 & 10 & 50 & 100 \\
7 & 20\% & 30\% & 37\% & 41\% \\
8 & 18\% & 29\% & 34\% & 39\% \\
9 & 15\% & 26\% & 36\% & 36\%  \\
10 & 20\% & 26\% & 38\% & \textbf{45\%} \\
}
\captionof{table}{\t{FF} model performance on FF benchmarks}\label{tab:ff2}
\end{table}

In this case, we observe that with 100 samples, the length-10 model is able to solve $45\%$ of the benchmarks. It surpasses the performance of Neural FlashFill~\cite{neuralflashfill}, which achieves an accuracy of 23\% with 100 samples and 34\% with 1000 samples. On further inspection of the benchmarks, we find that only $50\%$ of the benchmarks can be solved with programs of length $\leq 10$ in our DSL. If we normalize across this, we see that we can solve $\mathbf{90\%}$ of all solvable benchmarks. This indicates that our model is capable of learning to synthesize realistic programs.




\subsubsection{\t{FF++} Model}\label{ffpresults}

\paragraph{Baseline Performance with uniform search}
We first present the baseline results of uniform search. Since the DSL has expanded, the uniform search performs slightly worse and can only achieve an accuracy of about 11\% with 100 samples.
\begin{table}[h]
\centering
\pgfplotstabletypeset{
Samples & 10 & 50 & 100 \\
Performance & 3\% & 8\% & 11\% \\
}
\captionof{table}{Uniform search on FF benchmarks with full DSL}\label{tab:ffpu}
\end{table}

\paragraph{FlashFill benchmark performance}
The results for evaluating the \t{FF++} model on the FlashFill benchmarks is shown in Table~\ref{tab:ffp2}. The length 10 models can still remarkably solve 37\% of the benchmarks even with the extended DSL.

\begin{table}[h]
\centering
\pgfplotstabletypeset{
Length & 10 & 50 & 100 \\
7 & 28\% & 31\% & 33\% \\
8 & 24\% & 32\% & 34\% \\
9 & 19\% & 32\% & 34\%  \\
10 & 24\% & 34\% & \textbf{37\%} \\
}
\captionof{table}{\t{FF++} model performance on FF benchmarks}\label{tab:ffp2}
\end{table}

\section{Related Work}

We describe the related work from the domains of VSA-based programming by example systems and neural program induction and synthesis systems.

\paragraph{Programming By Example for String Manipulations}
There has been much recent work on designing version space algebra-based PBE systems for performing data transformation and extraction. FlashFill~\cite{flashfill,cacm12} is a PBE system that performs regular expression based string transformations using examples. Given an input-output example string, FlashFill first searches over all possible ways to decompose the output string and represent the set of those sub-programs concisely using a DAG data structure. This VSA-based approach has then been extended to also build PBE systems for number transformations~\cite{cav12}, table joins~\cite{vldb12}, data extraction~\cite{flashextract}, and data reshaping~\cite{flashrelate}. While these methods are interpretable and tractable, they are unscalable to any additions of new functionality. $\tool$, unlike the VSA-based PBE systems, is trained automatically using the R3NN network by sampling several thousands of programs from arbitrary DSLs.

%

\paragraph{Neural Program Induction and Synthesis}
There has been a plethora of recent work in both neural program induction and neural program synthesis. The goal in neural program induction is to teach neural networks the functional behavior of a program by augmenting the neural networks with additional computational modules such as Neural GPU~\cite{neuralgpu}, Neural Turing Machine~\cite{neuraltm}, and stacks-augmented RNNs~\cite{stackrnn}. One limitation of these architectures is that although they are able to learn the functional behavior, they do not expose an interpretable program back to the user. In addition, they need to be trained per task separately, representing a lack of strong generality. More recent work, such as Terpret~\cite{terpret} and Neural-RAM~\cite{neuralram} seek to mitigate the interpretability issue but they need to be trained for each individual benchmark problem, which is prohibitively expensive.

A recent approach was proposed to use the R3NN-based neural architectures to synthesize programs in a DSL similar to that of FlashFill~\cite{neuralflashfill}. We employ the same architecture but in a different DSL consisting of APIs at the core level of expressions. The APIs allows the program depth to be shallower than programs in a DSL with more primitives, and we investigate if that can make the task of automatically learning a search strategy easier for the R3NN. We argue that imposing higher-order functions is much more extensible and more akin to human-like programming.

\section{Future Work}
There are a number of ways that we can extend the results and techniques presented in this paper to yield both improvements in the current numbers as well as allow us to scale to larger programs.
\subsection{Function embeddings}
We rely on the R3NN and the input-output encoder to implicitly encode the semantics of each function, and we've shown through a number of experiments that the tree model is capable of doing so. This is impressive in its own right, but in order to improve the performance further, we should extend the model to support explicit, continuous representations of each function. This can be achieved in a number of ways - the simplest of which involves encoding each function as a randomly initialized vector and allowing the model to attend to API functions that may be relevant to the input-output examples. We can freeze the embeddings, or we can elect to backpropagate errors through both the attention mechanism and the embeddings, and jointly learn these representations. This represents a principled approach to adding new functions and method is easy to extend to additional API functions that we may choose to add.

\subsection{Divide and conquer}
Function embeddings allow us to perform better on existing problems by giving the model more information as to what choices to make when generating the tree. However, this does not resolve the issue of scalability. Even with function embeddings, as the inputs and outputs grow in size and complexity, we have no scalable method of performing inference over which programs to synthesize. However, instead of viewing the problem as a whole, we can break up the problem into smaller pieces and try to solve each subpiece and concatenate the answers together. This divide-and-conquer approach allows us to treat larger problems as conglomerations of a number of smaller problems. This procedure requires two general mechanisms: one module will need to predict how to split the output string into smaller, meaningful chunks, and the second module will consume each input-output piece, synthesize the correct program, and each piece will eventually be concatenated together. This is especially convenient in this problem setting because the FlashFill language is one that is focused on concatenations, so we lose no generality in being able to solve the problem.

\subsection{Extending the DSL}
An interesting extension of the DSL is to add multi-argument API function calls. This could yield more general API functions, such as \t{GetNthObj(n, o)}, and could replace functions like \t{GetFirstWord}, \t{GetSecondNumber}. In addition, we can also add multi-argument \t{Concat} functions; this idea goes neatly with the divide-and-conquer approach and can be used to help scale the model to synthesize larger programs.

\subsection{Batching Trees}
While the divide-and-conquer approach is an algorithmic improvement to speed up the process of training the model, we can also take advantage of the model to incorporate faster batching proocols. Using a tree-based generative models allows us to batch operations together that occur at the same depth in the tree because each operation is indepenedent of all of its siblings. Moreover, we can also batch multiple trees together for increased performance.

\section{Conclusion}

In this paper, we presented $\tool$, a system that tries to automatically learn a synthesis algorithm given a DSL. In particular, we designed a DSL consisting of APIs as first class constructs that allows the system to perform richer tasks using small sized programs. We used the recently introduced R3NN neural architecture to automatically learn a synthesis algorithm for our DSL. The preliminary results suggest that the system is able to efficiently learn programs up to size 10 with about 45\% accuracy on real-world benchmarks. We believe this direction of using neural architectures to automatically develop synthesis algorithms for PBE systems can lead to big advancements in program synthesis techniques and make it more generally applicable to many new domains.

\bibliographystyle{named}
\bibliography{references}


\onecolumn
\appendix
\begin{landscape}
\section*{Appendix}
\vspace{-0.15cm}
\section{The Complete Set of APIs} \label{apis}
\vspace{-0.25cm}
{\tiny
\begin{figure}[!htpb]
\bgroup
\def\arraystretch{0.9}
\begin{tabular}{|l|l|l|l|}
\hline
\multicolumn{1}{|l}{\tt \textbf{LookUp} (18)} & \multicolumn{1}{|l|}{\tt \textbf{Transform} (13)} & \multicolumn{2}{l|}{\tt \textbf{Regex} (104)} \\
\hline
\tt{ GetStreetNum } & \tt{ GetStateFromCity } & \tt{GetFirstWord} & \tt{GetFourthToLastNumber} \\
\hline
\tt{ GetStreetName } & \tt{ GetCityFromZipcode} & \tt{GetSecondWord} & \tt{GetFifthToLastNumber} \\
\hline
\tt{ GetAptNum } & \tt{GetStateAbbrFromState  } & \tt{GetThirdWord} & \tt{GetFirstAlpha} \\
\hline
\tt{ GetCityName } & \tt{ GetStateFromStateAbbr } & \tt{GetFourthWord} & \tt{GetSecondAlpha} \\
\hline
\tt{ GetStateName } & \tt{ GetFirstInitial } & \tt{GetFifthWord} & \tt{GetThirdAlpha}  \\
\hline
\tt{ GetStateAbbr } & \tt{ GetLastInitial } & \tt{GetLastWord} & \tt{GetFourthAlpha} \\
\hline
\tt{ GetZipcode } & \tt{ GetStockSymbolFromCEO } & \tt{GetSecondToLastWord} & \tt{GetFifthAlpha} \\
\hline
\tt{ GetFirstName } & \tt{ GetCEOFromCompany } & \tt{GetThirdToLastWord} & \tt{GetLastAlpha} \\
\hline
\tt{ GetLastName } & \tt{ GetCompanyFromStockSymbol } & \tt{GetFourthToLastWord} & \tt{GetSecondToLastAlpha} \\
\hline
\tt{ GetTitle } & \tt{ GetOrdinalFromDate } & \tt{GetFifthToLastWord} & \tt{GetThirdToLastAlpha} \\
\hline
\tt{ GetSuffix } & \tt{ GetMonthFromDate } & \tt{GetFirstNumber} & \tt{GetFourthToLastAlpha} \\
\hline
\tt{ GetCompany } & \tt{ GetWeekdayFromDate } & \tt{GetSecondNumber} & \tt{GetFifthToLastAlpha} \\
\hline
\tt{ GetCEO } & \tt{ GetYearFromDate } & \tt{GetThirdNumber} & \tt{GetFirstWS} \\
\hline
\tt{ GetStockSymbol } & \tt{  } & \tt{GetFourthNumber} & \tt{GetSecondWS} \\
\hline
\tt{ GetWeekday } & \tt{  } & \tt{GetFifthNumber} & \tt{GetThirdWS} \\
\hline
\tt{ GetMonth } & \tt{  } & \tt{GetLastNumber} & \tt{GetFourthWS} \\
\hline
\tt{ GetYear } & \tt{  } & \tt{GetSecondToLastNumber} & \tt{GetFifthWS} \\
\hline
\tt{ GetDate } & \tt{  } & \tt{GetThirdToLastNumber} & \tt{GetLastWS} \\
\hline
\end{tabular}
\egroup
\newline
\newline
\begin{tabular}{|l|l|l|l|}
\hline
\multicolumn{4}{|l|}{\tt \textbf{Regex} (cont.)} \\
\hline
\tt{GetSecondToLastWS} & \tt{GetFirstSpaceToEnd} & \tt{GetFirstTwoChar} & \tt{GetFifthToLastCapsWord} \\
\hline
\tt{GetThirdToLastWS} & \tt{GetStartToLastSpace} & \tt{GetFirstThreeChar} & \tt{GetFirstPropercaseWord} \\
\hline
\tt{GetFourthToLastWS} & \tt{GetLastSpaceToEnd} & \tt{GetFirstFourChar} & \tt{GetSecondPropercaseWord} \\
\hline
\tt{GetFifthToLastWS} & \tt{GetStartToDash} & \tt{GetFirstFiveChar} & \tt{GetThirdPropercaseWord} \\
\hline
\tt{TrimSpaces} & \tt{GetFirstDashToSecondDash} & \tt{GetFirstDigit} & \tt{GetFourthPropercaseWord} \\
\hline
\tt{TrimLeadingZeros} & \tt{GetLastDashToEnd} & \tt{GetFirstTwoDigit} & \tt{GetFifthPropercaseWord} \\
\hline
\tt{GetIdentity} & \tt{GetStartToFirstComma} & \tt{GetFirstThreeDigit} & \tt{GetAllPropercaseWords} \\
\hline
\tt{ReplaceSpacesWithDashes} & \tt{GetWordBetweenFirstAndSecondComma} & \tt{GetFirstFourDigit} & \tt{GetLastPropercaseWord} \\
\hline
\tt{ReplaceSpacesWithCommas} & \tt{GetWordBetweenSecondAndThirdComma} & \tt{GetFirstFiveDigit} & \tt{GetSecondToLastPropercaseWord} \\
\hline
\tt{ReplaceSpacesWithUnderscores} & \tt{GetLastCommaToEnd} & \tt{GetFirstCapsWord} & \tt{GetThirdToLastPropercaseWord} \\
\hline
\tt{ToLowercase} & \tt{GetWordBetweenCommaSpaceAndEnd} & \tt{GetSecondCapsWord} & \tt{GetFourthToLastPropercaseWord} \\
\hline
\tt{ToUppercase} & \tt{GetStartToParan} & \tt{GetThirdCapsWord} & \tt{GetFifthToLastPropercaseWord} \\
\hline
\tt{ToPropercase} & \tt{GetStartToFirstColon} & \tt{GetFourthCapsWord} & \tt{GetAllLetters} \\
\hline
\tt{GetWordBetweenStartAndAt} & \tt{GetStartToSecondColon} & \tt{GetFifthCapsWord} & \tt{GetAllNumbers} \\
\hline
\tt{GetWordBetweenAtAndEnd} & \tt{GetStringBetweenLastColonToEnd} & \tt{GetLastCapsWord} & \tt{ } \\
\hline
\tt{GetWordBetweenStartAndDot} & \tt{GetStringBetweenLastFirstAndSecondQuote} & \tt{GetSecondToLastCapsWord} & \tt{ } \\
\hline
\tt{GetWordBetweenDotAndEnd} & \tt{GetStartToEndOfFirstNumber} & \tt{GetThirdToLastCapsWord} & \tt{ } \\
\hline
\tt{GetStartToFirstSpace} & \tt{GetFirstChar} & \tt{GetFourthToLastCapsWord} & \tt{ } \\
\hline
\end{tabular}
\caption{Full list of functions used by our model; there are 18 lookup functions, 13 transform functions, and 104 regex functions. All functions take the input string as the input and produce a single output string, which can then be concatenated or nested with other function calls. Recall that we provide no function semantics of any of the above functions to the model; the model implicitly learns latent representations of each function. Therefore, any number of functions can be added to this library and the underlying learning algorithms will not need modifications.}
\end{figure}
}
\end{landscape}

\section{Samples of Training Data} \label{ffp_samples}

 \begin{figure}[!htpb]

{\centering
\bgroup
\def\arraystretch{1.1}
\begin{tabular}{|p{0.4\linewidth}|p{0.25\linewidth}|p{0.25\linewidth}|}
\hline
\multicolumn{3}{ |p{0.9\textwidth}| } {\texttt{\textbf{DAPIP Prediction:} (Concat (ConstStr CONST10) (GetStreetName (arg inp)))}} \\
\hline
{\tt \textbf{Inputs:}} & {\tt \textbf{Outputs:}} & {\tt \textbf{Predictions:}} \\
\hline
{\tt \} summer Impulse St. Pellerin} & {\tt Mr.Impulse St. }  & \textcolor{ForestGreen} {\tt Mr.Impulse St. } \\
{\tt Hensley Bag St. HI Rinaldo Nolan @ } & {\tt Mr.Bag St. }  & \textcolor{ForestGreen} {\tt Mr.Bag St. } \\
{\tt hook Gertha \% Plate St. hobbies MT } & {\tt Mr.Plate St. }  & \textcolor{ForestGreen} {\tt Mr.Plate St. } \\
{\tt discussion Mcfarlin . Straw St. } & {\tt Mr.Straw St. }  & \textcolor{ForestGreen} {\tt Mr.Straw St. } \\
{\tt hobbies Anger St. Twitty Downing ? } & {\tt Mr.Anger St. }  & \textcolor{ForestGreen} {\tt Mr.Anger St. } \\
\hline
\end{tabular}
\egroup
}
\vspace{0.6cm}

{\centering
\bgroup
\def\arraystretch{1.1}
\begin{tabular}{|p{0.4\linewidth}|p{0.25\linewidth}|p{0.25\linewidth}|}
\hline
\multicolumn{3}{ |p{0.9\textwidth}| } {\texttt{\textbf{DAPIP Prediction:} (Concat (ConstStr CONST36) (GetStateName (arg inp)))}} \\
\hline
{\tt \textbf{Inputs:}} & {\tt \textbf{Outputs:}} & {\tt \textbf{Predictions:}} \\
\hline
{\tt MA , North Carolina Zehr Gilma } & {\tt ~North Carolina }  & \textcolor{ForestGreen} {\tt  ~North Carolina } \\
{\tt Utah Evelia \% Nancy } & {\tt ~Utah}  & \textcolor{ForestGreen} {\tt ~Utah } \\
{\tt Josh skin . Missouri Agudelo } & {\tt ~Missouri }  & \textcolor{ForestGreen} {\tt ~Missouri } \\
{\tt yarn drawer ` Indiana } & {\tt ~Indiana }  & \textcolor{ForestGreen} {\tt ~Indiana } \\
{\tt Sandidge ) key Indiana } & {\tt ~Indiana }  & \textcolor{ForestGreen} {\tt ~Indiana } \\
\hline
\end{tabular}
\egroup
}
\vspace{0.6cm}

{\centering
\bgroup
\def\arraystretch{1.1}
\begin{tabular}{|p{0.4\linewidth}|p{0.25\linewidth}|p{0.25\linewidth}|}
\hline
\multicolumn{3}{ |p{0.9\textwidth}| } {\texttt{\textbf{DAPIP Prediction:} (Concat (GetStateAbbrFromState (arg inp)) (ConstStr CONST25))}} \\
\hline
{\tt \textbf{Inputs:}} & {\tt \textbf{Outputs:}} & {\tt \textbf{Predictions:}} \\
\hline
{\tt Elza Foot Locker Illinois @bo.com Mollett } & {\tt IL * }  & \textcolor{ForestGreen} {\tt IL * } \\
{\tt \$ can Sound St. mist Nevada } & {\tt NV * }  & \textcolor{ForestGreen} {\tt NV * } \\
{\tt Harpin Utah . Reali RI Laurinda Borden } & {\tt UT * }  & \textcolor{ForestGreen} {\tt UT * } \\
{\tt ) Connecticut Belt Mortimer } & {\tt CT * }  & \textcolor{ForestGreen} {\tt  CT * } \\
{\tt Danita ~ Tennessee throat } & {\tt TN * }  & \textcolor{ForestGreen} {\tt TN * } \\
\hline
\end{tabular}
\egroup
}
\vspace{0.6cm}

{\centering
\bgroup
\def\arraystretch{1.1}
\begin{tabular}{|p{0.4\linewidth}|p{0.25\linewidth}|p{0.25\linewidth}|}
\hline
\multicolumn{3}{ |p{0.9\textwidth}| } {\texttt{\textbf{DAPIP Prediction:} (GetSecondToLastWS (arg (GetCEO (arg inp))))}} \\
\hline
{\tt \textbf{Inputs:}} & {\tt \textbf{Outputs:}} & {\tt \textbf{Predictions:}} \\
\hline
{\tt Eldora John Thain Marotta } & {\tt John }  & \textcolor{ForestGreen} {\tt John } \\
{\tt Marya clover Sundar Pichai } & {\tt Sundar }  & \textcolor{ForestGreen} {\tt Sundar } \\
{\tt 327 drawer Gregory Wasson Kristian } & {\tt Gregory }  & \textcolor{ForestGreen} {\tt Gregory } \\
{\tt ! AOL Inc. Rinaldo quicksand James Gorman } & {\tt James }  & \textcolor{ForestGreen} {\tt James } \\
{\tt Richard Johnson Barbie Gasaway } & {\tt Richard }  & \textcolor{ForestGreen} {\tt Richard } \\
\hline
\end{tabular}
\egroup
}
\vspace{0.6cm}

\caption{Selected samples of correct model predictions on the FlashFill++ synthetic test set. These samples also provide samples of the nature of the programmatically generated training data for the FF and FF++ models. Note that because these programs rely on the semantic Lookup and Transform APIs, there is no provided reference FlashFill program, as there would be no program that could solve these synthetic benchmarks.}
\label{fig:ffp_samples}
\end{figure}

\newpage
\section{Samples of Solved FlashFill Benchmarks}\label{ff_samples}

 \begin{figure}[!htpb]

{\centering
\bgroup
\def\arraystretch{1.1}
\begin{tabular}{|p{0.3\linewidth}|p{0.3\linewidth}|p{0.3\linewidth}|}
\hline
\multicolumn{3}{ |p{0.9\textwidth}| } {\texttt{\textbf{FlashFill Program:} (SubStr (RegPos (RegexStr (ConstStr "0-0")) (k 1) (dir End))(RegPos (RegexStr REGEX4) (k 4) (dir End)))}} \\
\hline
\multicolumn{3}{ |p{0.9\textwidth}| } {\texttt{\textbf{DAPIP Prediction:} (TrimLeadingZeros (arg (GetFirstDashToSecondDash (arg Inp))))}} \\
\hline
{\tt \textbf{Inputs:}} & {\tt \textbf{Outputs:}} & {\tt \textbf{Predictions:}} \\
\hline
{\tt 09:40-09:50 } & {\tt 9:50 }  & \textcolor{ForestGreen} {\tt 9:50 } \\
{\tt 09:50-08:30 } & {\tt 8:30 }  & \textcolor{ForestGreen} {\tt 8:30 } \\
{\tt 09:50-07:30 } & {\tt 7:30 }  & \textcolor{ForestGreen} {\tt 7:30 } \\
{\tt 09:50-09:55 } & {\tt 9:55 }  & \textcolor{ForestGreen} {\tt 9:55 } \\
{\tt 05:50-06:30 } & {\tt 6:30 }  & \textcolor{ForestGreen} {\tt 6:30 } \\
\hline
\end{tabular}
\egroup
}
\vspace{0.6cm}

{\centering
\bgroup
\def\arraystretch{1.1}
\begin{tabular}{|p{0.3\linewidth}|p{0.3\linewidth}|p{0.3\linewidth}|}
\hline
\multicolumn{3}{ |p{0.9\textwidth}| } {\texttt{\textbf{FlashFill Program:} (Concat (SubStr (RegPos (RegexStr REGEX8) (k 1) (dir End)) (RegPos (RegexStr REGEX1) (k 1) (dir End))) (ConstStr "@"))}} \\
\hline
\multicolumn{3}{ |p{0.9\textwidth}| } {\texttt{\textbf{DAPIP Prediction:} (Concat (ToLowercase (arg (GetFirstWord (arg Inp))) (ConstStr CONST13))}} \\
\hline
{\tt \textbf{Inputs:}} & {\tt \textbf{Outputs:}} & {\tt \textbf{Predictions:}} \\
\hline
{\tt Sophia Underwood } & {\tt sophia@ }  & \textcolor{ForestGreen} {\tt sophia@ } \\
{\tt Logan Smith } & {\tt logan@ }  & \textcolor{ForestGreen} {\tt logan@ } \\
{\tt Lucas Janckle } & {\tt lucas@ }  & \textcolor{ForestGreen} {\tt lucas@ } \\
{\tt Audrey Bennette } & {\tt audrey@ }  & \textcolor{ForestGreen} {\tt audrey@ } \\
{\tt Amelia Ford } & {\tt amelia@ }  & \textcolor{ForestGreen} {\tt amelia@ } \\
\hline
\end{tabular}
\egroup
}
\vspace{0.6cm}

{\centering
\bgroup
\def\arraystretch{1.1}
\begin{tabular}{|p{0.3\linewidth}|p{0.3\linewidth}|p{0.3\linewidth}|}
\hline
\multicolumn{3}{ |p{0.9\textwidth}| } {\texttt{\textbf{FlashFill Program:} (Concat (SubStr (RegPos (RegexStr REGEX8) (k 1) (dir End)) (RegPos (RegexStr REGEX4) (k 1) (dir End))) (ConstStr "]"))}} \\
\hline
\multicolumn{3}{ |p{0.9\textwidth}| } {\texttt{\textbf{DAPIP Prediction:} (Concat (GetStartToEndOfFirstNumber (arg (ToUppercase (arg Inp))) (ConstStr CONST12))}} \\
\hline
{\tt \textbf{Inputs:}} & {\tt \textbf{Outputs:}} & {\tt \textbf{Predictions:}} \\
\hline
{\tt [CPT-00350 } & {\tt [CPT-00350] }  & \textcolor{ForestGreen} {\tt [CPT-00350] } \\
{\tt [CPT-11523] } & {\tt [CPT-11523] }  & \textcolor{ForestGreen} {\tt [CPT-11523] } \\
{\tt [CPT-23412] } & {\tt [CPT-23412] }  & \textcolor{ForestGreen} {\tt [CPT-23412] } \\
{\tt [CPT-23412 } & {\tt [CPT-23412] }  & \textcolor{ForestGreen} {\tt  [CPT-23412] } \\
{\tt [CPT-2422] } & {\tt [CPT-2422] }  & \textcolor{ForestGreen} {\tt [CPT-2422] } \\
\hline
\end{tabular}
\egroup
}
\vspace{0.6cm}

{\centering
\bgroup
\def\arraystretch{1.1}
\begin{tabular}{|p{0.3\linewidth}|p{0.3\linewidth}|p{0.3\linewidth}|}
\hline
\multicolumn{3}{ |p{0.9\textwidth}| } {\texttt{\textbf{FlashFill Program:} (SubStr (RegPos (RegexStr REGEX8) (k 1) (dir End)) (RegPos (RegexStr REGEX4) (k 1) (dir End)))}} \\
\hline
\multicolumn{3}{ |p{0.9\textwidth}| } {\texttt{\textbf{DAPIP Prediction:} (GetLastNumber (arg (TrimSpaces (arg GetFirstAlpha (arg Inp)))))}} \\
\hline
{\tt \textbf{Inputs:}} & {\tt \textbf{Outputs:}} & {\tt \textbf{Predictions:}} \\
\hline
{\tt 1:42:00 AM } & {\tt 1 }  & \textcolor{ForestGreen} {\tt 1 } \\
{\tt 4:18:00 AM } & {\tt 4 }  & \textcolor{ForestGreen} {\tt 4 } \\
{\tt 6:54:00 PM } & {\tt 6 }  & \textcolor{ForestGreen} {\tt 6 } \\
{\tt 11:06:00 AM } & {\tt 11 }  & \textcolor{ForestGreen} {\tt 11 } \\
{\tt 9:12:00 AM } & {\tt 9 }  & \textcolor{ForestGreen} {\tt 9 } \\
\hline
\end{tabular}
\egroup
}
\vspace{0.6cm}

\caption{Selected samples of correct model predictions on the Flashfill test set. We additionally provide the program that FlashFill inferred given the input-output pairs, and contrast that with DAPIP's prediction. Note that DAPIP programs have a much higher level of expressivity and interpretability.}
\label{fig:ff_samples}
\end{figure}

\newpage
\section{Samples of Unsolved FlashFill Benchmarks} \label{ff_unsolved}

 \begin{figure}[!htpb]

{\centering
\bgroup
\def\arraystretch{1.1}
\begin{tabular}{|p{0.45\linewidth}|p{0.45\linewidth}|}
\hline
\multicolumn{2}{ |p{0.9\textwidth}| } {\texttt{\textbf{FlashFill Program:} (Concat (Concat (Concat (Concat (Concat (Concat (Concat (Concat (Concat (Concat (SubStr (RegPos (RegexStr REGEX8) (k 1) (dir End)) (RegPos (RegexStr REGEX1) (k 1) (dir End))) (ConstStr ",")) (SubStr (RegPos (RegexStr REGEX7) (k 1) (dir End)) (RegPos (RegexStr REGEX1) (k 2) (dir End)))) (ConstStr ",")) (SubStr (RegPos (RegexStr REGEX7) (k 2) (dir End)) (RegPos (RegexStr REGEX1) (k 3) (dir End)))) (ConstStr ",")) (SubStr (RegPos (RegexStr REGEX7) (k 3) (dir End)) (RegPos (RegexStr REGEX1) (k 4) (dir End)))) (ConstStr ".")) (ConstStr "and")) (ConstStr ".")) (SubStr (RegPos (RegexStr REGEX7) (k 4) (dir End)) (RegPos (RegexStr REGEX10) (k 1) (dir End))))}} \\
\hline
{\tt \textbf{Inputs:}} & {\tt \textbf{Outputs:}} \\
\hline
{\tt Tom Mickey Minnie Donald Daffy } & {\tt Tom,Mickey,Minnie,Donald.and.Daffy } \\
{\tt Ben Bill Jerry Meyer Rahul } & {\tt Ben,Bill,Jerry,Meyer.and.Rahul } \\
{\tt Shahrukh Aamir Salman Amitabh Ajay } & {\tt Shahrukh,Aamir,Salman,Amitabh.and.Ajay } \\
{\tt Kobe Lebron Dwayne Chris Kevin } & {\tt Kobe,Lebron,Dwayne,Chris.and.Kevin } \\
{\tt Earth Fire Wind Water Sun } & {\tt Earth,Fire,Wind,Water.and.Sun } \\
\hline
\end{tabular}
\egroup
}
\vspace{0.6cm}

{\centering
\bgroup
\def\arraystretch{1.1}
\begin{tabular}{|p{0.45\linewidth}|p{0.45\linewidth}|}
\hline
\multicolumn{2}{ |p{0.9\textwidth}| } {\texttt{\textbf{FlashFill Program:} (Concat (Concat (Concat (SubStr (RegPos (RegexStr REGEX8) (k 1) (dir End)) (RegPos (RegexStr REGEX4) (k 1) (dir End))) (SubStr (RegPos (RegexStr (ConstStr "1-")) (k 1) (dir End)) (RegPos (RegexStr REGEX4) (k 2) (dir End)))) (SubStr (RegPos (RegexStr (ConstStr "-")) (k 2) (dir End)) (RegPos (RegexStr REGEX4) (k 3) (dir End)))) (SubStr (RegPos (RegexStr (ConstStr "-")) (k 3) (dir End)) (RegPos (RegexStr REGEX4) (k 4) (dir End))))}} \\
\hline
{\tt \textbf{Inputs:}} & {\tt \textbf{Outputs:}} \\
\hline
{\tt 1-452-789-4567 } & {\tt 14527894567 } \\
{\tt 1-503-897-4567 } & {\tt 15038974567 } \\
{\tt 1-408-789-4561 } & {\tt 14087894561 } \\
{\tt 1-406-789-1562 } & {\tt 14067891562 } \\
{\tt 1-845-456-7891 } & {\tt 18454567891 } \\
\hline
\end{tabular}
\egroup
}
\vspace{0.6cm}

\caption{Selected samples from the FlashFill benchmarks that could not be solved by our model; benchmarks such as these constitute the 55\% of FlashFill benchmarks that our model cannot solve. At present, the maximum length of programs that DAPIP can produce is only 10, and these particular benchmarks would require much longer programs. However, if the batching of trees can be done more efficiently, the system can be trained on longer programs and it is conceivable that these benchmarks can be solved. }
\label{fig:ff_unsolved}
\end{figure}

\end{document}